\documentclass[10pt,twocolumn,letterpaper]{article}

\usepackage{cvpr}      
\usepackage{multirow}

\definecolor{cvprblue}{rgb}{0.21,0.49,0.74}
\usepackage[pagebackref,breaklinks,colorlinks,allcolors=cvprblue]{hyperref}

\def\confName{CVPR}
\def\confYear{2026}

\title{LEMMA: Laplacian pyramids for Efficient Marine SeMAntic Segmentation}

\author{
Ishaan Gakhar,\thanks{Equal contribution.} \quad
Laven Srivastava,\footnotemark[1] \quad
Sankarshanaa Sagaram, \quad
Aditya Kasliwal, \quad
Ujjwal Verma,\\
Manipal Institute of Technology, Manipal Academy of Higher Education, India\\
{\tt\small ishaangakhar04@gmail.com \quad lavensri@gmail.com}
}

\begin{document}
\maketitle
\begin{abstract}
Semantic segmentation in marine environments is crucial for the autonomous navigation of unmanned surface vessels (USVs) and coastal Earth Observation events such as oil spills. However, existing methods, often relying on deep CNNs and transformer-based architectures, face challenges in deployment due to their high computational costs and resource-intensive nature. These limitations hinder the practicality of real-time, low-cost applications in real-world marine settings.

To address this, we propose LEMMA, a lightweight semantic segmentation model designed specifically for accurate remote sensing segmentation under resource constraints. The proposed architecture leverages Laplacian Pyramids to enhance edge recognition, a critical component for effective feature extraction in complex marine environments for disaster response, environmental surveillance, and coastal monitoring. By integrating edge information early in the feature extraction process, LEMMA eliminates the need for computationally expensive feature map computations in deeper network layers, drastically reducing model size, complexity and inference time. LEMMA demonstrates state-of-the-art performance across datasets captured from diverse platforms while reducing trainable parameters and computational requirements by up to 71x, GFLOPs by up to 88.5\%, and inference time by up to 84.65\%, as compared to existing models. Experimental results highlight its effectiveness and real-world applicability, including 93.42\% IoU on the Oil Spill dataset and 98.97\% mIoU on Mastr1325. 
\end{abstract}    
\section{Introduction}
\label{sec:intro}

Earth observation (EO) from Unmanned Aerial Vehicles (UAV) has become a cornerstone of large-scale environmental monitoring, enabling persistent surveillance of coastal ecosystems, maritime infrastructure, and pollution events such as oil spills and chemical discharges \cite{temitope2020advances} \cite{miao2018automatic}. High-resolution UAV-based EO imagery is increasingly used for rapid disaster response, coastal zone management, and environmental impact assessment due to its low deployment cost, flexible coverage, and high spatial resolution compared to satellite-based sensing \cite{miao2018automatic} \cite{taipalmaa2019high}. However, EO segmentation in maritime and coastal regions remains particularly challenging because of strong specular reflections, low inter-class contrast between water and thin surface films, atmospheric illumination variations, and dynamic surface textures caused by waves and wind \cite{temitope2020advances} \cite{zhou2019underwater}. These challenges impose strict requirements on segmentation models to be not only accurate but also computationally efficient for deployment on resource-constrained aerial platforms used in real-world EO missions. Consequently, there is a growing demand for lightweight, edge-aware semantic segmentation frameworks that can deliver reliable geospatial delineation in complex EO scenarios while maintaining real-time performance.

Semantic segmentation is a cornerstone task in computer vision, enabling pixel-level understanding for various real-world scenarios, and finds applications in UAV domains, autonomous driving systems \cite{thisanke2023semantic}, robotic vision \cite{manakitsa2024review}, unmanned surface vehicles (USVs), and oil spill identification from aerial drone imagery. These tasks are relevant to operational hazards and environmental damage \cite{long2015fully}, \cite{fiscella2000oil}. Traditional state-of-the-art (SOTA) segmentation models are typically resource-intensive \cite{bovcon2022wasr}, requiring significant memory and computational power, making them impractical for deployment on edge devices such as USVs or drones. The growing demand for real-time, on-device processing has thus driven the need for lightweight, efficient models capable of delivering high performance in resource-constrained environments \cite{xie2015holistically}.

\begin{figure*}[]
    \centering
    \includegraphics[width=\textwidth]{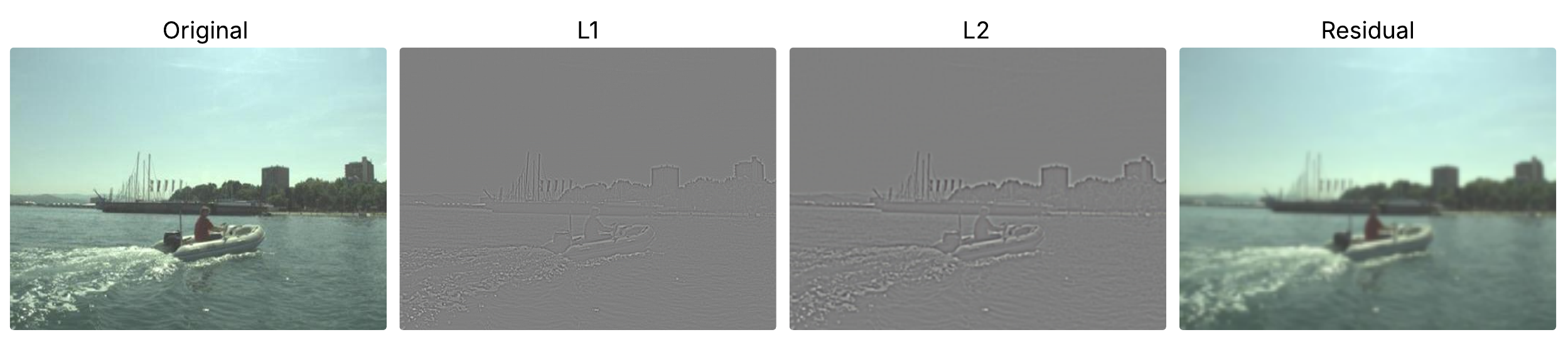}
    \caption{Visualization of the Laplacian pyramid of depth 2 of an RGB image. Vital spatial and structural information is present at each level, notably in the edges available at different resolutions. Here, $L_1$, $L_2$ and Residual ($L_3$) refer to the first, second and last layer of the decomposed Laplacian Pyramid.}
    \label{fig:laplacian_vis}
\end{figure*}

The domain of marine semantic segmentation has evolved through diverse architectural paradigms, encompassing both general-purpose and domain-specific approaches. Early works adapted terrestrial segmentation networks like FCN \cite{long2015fully}, and SegNet \cite{badrinarayanan2017segnet} to maritime environments with limited success. More sophisticated architectures emerged with DeepLabv3 \cite{chen2017rethinking} and PSPNet \cite{zhao2017pyramid}, leveraging atrous convolutions and pyramid pooling for improved contextual understanding. Maritime-specific architectures such as WaSR \cite{bovcon2022wasr} and WaSR-T incorporated temporal consistency and transformer modules, while MUNet \cite{abdollahi2022automatic} introduced multi-level attention mechanisms for maritime obstacle detection. 
The proposed methodology diverges from these approaches by employing a Laplacian pyramid decomposition chain with residual blocks at each level for efficient feature extraction, drastically reducing the computational footprint with up to 71 times fewer parameters while maintaining competitive performance across diverse maritime datasets from both surface and aerial perspectives.

Our motivation in employing Laplacian pyramids lies in the edge information returned in a one-shot manner by the decomposition of the pyramid. As visible in Fig. \ref{fig:laplacian_vis}, the edges of the obstacles (in this example) are visible at various resolutions and are easy to identify in the layers of the pyramid. This edge information allows for multi-level feature extraction, leading to superior results across images from diverse capturing platforms, like drone and USV data, at a much lower computational cost. This reduction in computational complexity enables the model to run efficiently on edge devices without compromising accuracy, making it ideal for real-world applications. The lightweight design of LEMMA bridges the gap between computational constraints and the need for precise segmentation in critical scenarios \cite{lee2023semantic}. Our key contributions are as follows.

\begin{itemize}
    \item Adapt the Laplacian Pyramid to extract vital edge information, enabling efficient and precise marine semantic segmentation.
    \item Validate the proposed methodology on two challenging and diverse tasks - obstacle segmentation in USV data and Oil Spill segmentation captured from aerial drones - demonstrating its broad applicability in real-world scenarios and deployment.
    \item Demonstrate SOTA results while reducing model parameters by up to 71x, GFLOPs by up to 88.5\%, and inference time by up to 84.65\%, showcasing suitability for deployment on resource-constrained platforms like drones, UAVs, and USVs.
\end{itemize}

Unlike prior applications of Laplacian pyramids, which focus on medical or terrestrial contexts, our work adapts the decomposition mechanism for low-cost, edge-aware segmentation in marine environments. We introduce a three-branch residual framework specifically designed to work with pyramid-level edge cues, enhancing thin-boundary prediction without heavy post-processing.

\section{Related Work}
\label{sec: related_works}
\begin{figure*}[t]
    \centering
    \includegraphics[width=\textwidth]{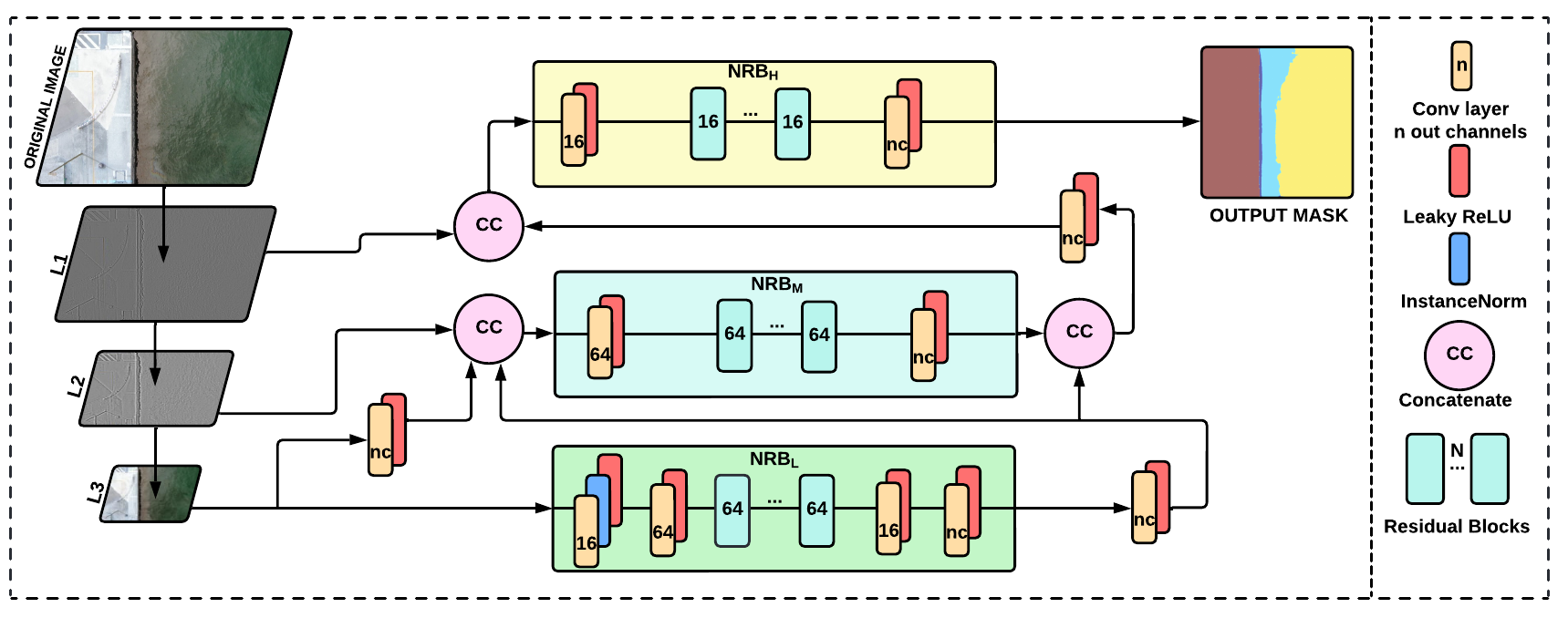}
    \caption{A schematic overview of the proposed model-LEMMA. The sections highlighted in green, blue, and yellow symbolize the LFB, MFB, and HFB, respectively. $L_1$, $L_2$, and Residual ($L_3$) represent the three layers of the decomposed Laplacian pyramid. The 'nc' for each dataset is the total number of classes in the dataset. Each residual block chain has the corresponding number of blocks, as explained in Section \ref{sec: Method}.}
    \label{fig:model_diagram}
\end{figure*}
In this study, previous works relevant to Laplacian pyramids and Marine segmentation have been summarised. Laplacian pyramids are a traditional computer vision technique which finds its applications in various tasks, like segmentation \cite{10.1007/978-3-319-46487-9_32}, image generation \cite{denton2015deep}, super-resolution \cite{kasliwal2024lapgsr}, style transfer \cite{lin2021drafting}, etc. 

\subsection{Laplacian Pyramid-Based Feature Extraction for Semantic Segmentation}

Laplacian pyramids have been widely employed in computer vision for multi-scale feature extraction due to their capability of preserving fine-grained edge details while enabling hierarchical image analysis. These methods are particularly effective in enhancing segmentation performance by capturing both low-frequency structural information and high-frequency edge details, which are essential for distinguishing fine boundaries in complex scenes. 

The concept of Laplacian pyramid decomposition was first introduced as a multi-resolution analysis method that decomposes an image into progressively lower-resolution representations while maintaining high-frequency details \cite{burt1987laplacian}. This hierarchical approach allows for efficient feature extraction and reconstruction, making it highly effective for segmentation tasks where both local and global contextual information are critical. Paris et al. \cite{paris2011local} extended this concept by introducing Local Laplacian Filters, which enhance contrast and detail preservation using Laplacian pyramids in an edge-aware manner. Their method allows for fine-scale control of contrast enhancement without introducing halo artifacts, making it suitable for edge-sensitive applications like semantic segmentation.



\subsection{Marine Segmentation}

Marine semantic segmentation as a computer vision task involves pixel-level classification of underwater imagery to distinguish marine objects, organisms, and terrain features from their surroundings \cite{wang2022underwater}. Recent advances in deep learning have introduced several state-of-the-art (SOTA) models designed specifically for the challenges of marine perception, such as varying water conditions, occlusions, and real-time processing constraints.

WaSRNet \cite{bovcon2022wasr} is a maritime segmentation network designed to handle water variability and false positives in USV navigation. It employs an encoder-decoder architecture with feature fusion mechanisms to improve segmentation in cluttered water scenes. By incorporating spatial attention and multi-level feature aggregation, WaSRNet effectively reduces false positives in wave-heavy environments, making it highly effective for maritime navigation in dynamic conditions. However, its encoder-decoder design requires 71.4 million parameters, limiting deployment on edge devices.

Building upon this, BEMSNet \cite{zhang2024bems} enhances boundary segmentation by introducing boundary-aware loss functions and edge-refining modules, achieving SOTA performance on various datasets. While it improves contour accuracy for thin structures like buoys and ship railings, the added boundary enhancement modules increase computational complexity, maintaining a parameter count of 71.4 million. This underscores a critical trade-off between precision and efficiency in safety-critical applications.

A Lightweight Dual-Branch Semantic Segmentation Network for Ship Navigation \cite{feng2024dual} proposes separate pathways for low-level and high-level feature extraction, reducing computational overhead compared to single-stream architectures. While this approach achieves real-time processing with 31 million parameters, its dual-branch design still struggles to balance accuracy and speed in highly dynamic marine environments with rapid lighting and surface variations.

The marine segmentation landscape reveals a consistent pattern: Although specialised domain methods achieve high accuracy, none effectively addresses the dual requirements of maintaining precision while enabling real-time deployment on resource-constrained marine platforms. Traditional approaches require tens of millions of parameters and hundreds of GFLOPS, creating a substantial computational barrier for practical marine deployment.
\section{Methodology}
\label{sec: Method}

This section details the architecture of the proposed methodology, LEMMA, a novel edge-recognition-based lightweight model employing Laplacian pyramids to bypass heavy feature map computation for Marine semantic segmentation. Drawing upon established research, we incorporate multi-scale processing with residual connections and feature concatenation across different levels as fundamental components of our design. As depicted in Fig. \ref{fig:model_diagram}, the original image is decomposed into a Laplacian pyramid of depth 3, resulting in three layers referred to as $L_1$, $L_2$ and $L_3$. Features are extracted from these layers at various scales, which undergo further processing in the High-level Feature Branch, Middle-level Feature Branch and Low-level Feature Branch, respectively. 

\subsection{Low-level Feature Branch}

The input to the Low-level Feature Branch (LFB) is $L_3$. Before being processed by the LFB, $L_3$ undergoes a convolution operation, followed by a leaky ReLU activation to be concatenated with $L_2$, before $L_2$ is processed by the Middle-level Feature Branch.

The LFB consists of various Convolution layers (conv), InstanceNorm, and Leaky ReLU activations. A chain of residual blocks is placed between these modules. These residual blocks consist of a conv layer, leaky ReLU, and another conv layer with a residual connection. The kernel size of 3 and the stride of 1 are constant throughout the architecture. NRB$_L$ denotes the number of these residual blocks. The conv layers before the residual block chain bring the number of channels of the feature map to 64 channels. By integrating the low-level details extracted by the LTB with the higher-level features, the network can construct a more comprehensive feature representation, setting the stage for advanced processing in subsequent network stages. 

Hence, the feature extraction can be expressed as:

\begin{equation}
    \begin{aligned}
    L_{3f} &= \text{LFB}(L_{3}) \\
    L_{3}' &= \text{LeakyReLU}(\text{TransposeConv}(L_3))
    \end{aligned}
\end{equation}

Additionally, before $L_3$ is processed by the LFB, it is upsampled using a transpose convolution and leaky ReLU activation ($L_{3}'$) and concatenated with $L_2$ and $L_{3f}$. This is implemented to retain the raw information present at this scale of the pyramid.

\subsection{Middle-level Feature Branch}

The Middle-level Feature Branch (MFB) receives the concatenation of $L_{3f}$, $L_{3}'$ and $L_2$. This feature map is then processed by residual blocks between convolution and leaky ReLU layers. NRB$_{M}$ indicates the number of these residual blocks. 

The MFB is responsible for refining and distilling key features received from $L_2$ and $L_3$. The accentuated edge information and fine details derived from Laplacian pyramids ($L_2$) help to bypass heavy feature map computation. These derived details allow the model to avoid an excessive number of parameters. The MFB extracts spatial and structural information by processing and
refining the joint feature maps of the LFB and the $L_2$
layer of the pyramid. This advanced feature representation is
crucial for the network’s subsequent stages, where such high-
level conceptualisations of the image are vital for advanced
visual tasks. 

The output of the conv and Leaky ReLU module placed after the residual blocks is further concatenated with the output of the LFB. This feature map now contains information refined from the Middle and Lower branches, thereby retaining structural awareness at different scales. This is subsequently passed through another conv and leaky ReLU module and concatenated with $L_3$ to finally be processed by the High-level Feature Branch.

Hence, the processing in the branch can be formulated as:

\begin{equation}
    \begin{aligned}
    L_{\text{concat}} &= [L_2; L_{3f}; L_3'] \\
    L_{\text{MFB}} &= \text{MFB}(L_{\text{concat}}) \\
    L_{\text{CC}} &= [L_{\text{MFB}}; L_{3f}] \\
    L_{\text{out}} &= \text{LeakyReLU}(\text{TransposeConv}(L_{\text{CC}}))
    \end{aligned}
\end{equation}

where $[;]$ represents concatenation.

\subsection{High-level Feature Branch}

The High-level Feature Branch (HFB) carries out further refinement and considers the intricacies present in $L_3$, the highest-resolution Laplacian layer. The information extracted from $L_3$, along with the merged features of lower branches, allows for accurate reconstruction, taking into account spatial and structural features, enabling swift reconstruction. The HFB starts with a conv layer and leaky ReLU, followed up by NRB$_{H}$ number of residual blocks. Finally, a conv layer and leaky ReLU activation follow to generate the mask. Following a structure similar to the MFB and LFB but only processing features of 16 channels instead of 64 channels, the HFB ensures lower GFLOPs while reconstructing a highly accurate mask swiftly and inexpensively. The feature processing is formally represented as:

\begin{equation}
\begin{aligned}
L_{\text{CC}} &= [L_1; L_{\text{out}}] \\
M_{\text{final}} &= \text{HFB}(L_{\text{CC}})
\end{aligned}
\end{equation}

where $[;]$ represents concatenation, and $M_{final}$ is the final multiclass segmentation mask.
\section{Experimental Setup and Results}  
\label{sec:experiments}

\begin{table*}[t]
\renewcommand{\arraystretch}{1.3}
\centering
\small
\begin{tabular}{|l|c|c|c|c|c}
\textbf{Model} & \textbf{mIoU} & \textbf{\#Params (M)} & \textbf{GFLOPs} & \textbf{Inference Time (ms)} \\ \hline
DeepLabv3 + Xception \cite{chen2018encoder} & 80.65 & 54.71 & 124.65 & 29.83 \\
ERFNet \cite{romera2017erfnet} & 81.68 & 2.06 & 22.16 & 15.07 \\
Pretrained SegNet \cite{badrinarayanan2017segnet} & 81.80 & 29.45 & 241.29 & 22.31 \\
CFPNet \cite{lou2021cfpnet} & 82.28 & \textbf{0.54} & \textbf{5.85} & 35.61 \\
ERFNetv2 \cite{mehta2019espnetv2} & 82.45 & 1.23 & 4.18 & 18.33 \\
DABNet \cite{li2019dabnet} & 82.64 & 0.75 & 7.76 & 15.59 \\
LETNet \cite{xu2023lightweight} & 83.18 & 0.94 & 10.63 & 73.15 \\
UISSNet + ResNet50 \cite{he2024uiss} & 83.56 & 75.42 & 256.59 & 35.15 \\
UISSNet + ResNet18 \cite{he2024uiss} & 83.91 & 44.11 & 195.15 & 24.94 \\
HRNet-w18 \cite{chen2022mobile} & 84.11 & 29.54 & 67.90 & 33.85 \\
LGCNet \cite{liu2025lgcgnet} & 84.14 & 0.69 & 9.05 & 28.67 \\
Pretrained UNet \cite{ronneberger2015u} + VGG16 \cite{simonyan2014very} & 86.01 & 24.89 & 339.01 & 48.29 \\
WODIS \cite{chen2021wodis} & 91.30 & 89.50 & - & 23.15 \\
PSPNet \cite{zhao2017pyramid} & 93.46 & 66.0 & 8.91  & 5.94 \\
Fast SCNN \cite{poudel2019fast} & 93.50 & 1.36 & - & 14.82 \\
DDRNet-s \cite{pan2022deep} & 94.50 & 17.05 & - & 12.66 \\
Segmenter \cite{strudel2021segmenter} & 94.80 & - & - & 18.73 \\
LDANet \cite{dai2025ldanet} & 96.16 & 18.53 & - & 12.35 \\
LightWeight \cite{taipalmaa2019high} & 96.81 & - & - & - \\
DeepLabv3 \cite{chen2017rethinking} & 97.67 & 48.0 & 123.11 & 47.55 \\
HRNet \cite{sun2019deep} & 97.87 & 63.0 & - & - \\
ShorelineNet \cite{yao2021shorelinenet} & 98.75 & 6.5 & - & - \\
SGAF \cite{chen2022water} & 98.88 & 63.0 & - & - \\
WaSR-T \cite{zust2022temporal} & 99.80 & 71.4 & 133.80 & - \\
BEMRF-Net \cite{cao2024bemrf} & \textbf{99.91} & 71.40 & 156.0 & - \\
\hline
\textbf{LEMMA (Ours)} & \textbf{98.97} & \textbf{1.07} & \textbf{17.83} & \textbf{7.3} \\
\hline
\end{tabular}
\caption{Comparison of various models with LEMMA on \textbf{mIoU} (in percent) across MaSTr1325 \cite{bovcon2019mastr} and Oil Spill Drone \cite{sels2024oilspill} datasets. Parameters, GFLOPs, and inference time are provided for both ResNet-50 and ResNet-101 backbones where applicable. Results for LEMMA are of residual block (Section \ref{sec: Method}) configurations 7,7,1 and 6,7,4 respectively (separated by forward slashes for Mastr1325 and Oil Spill). Values for SOTA comparisons separated by slashes refer to R50/R101 variants. Unreported values in original works or lack of publically available weights have been left blank.}
\label{tab:Mastr_results}
\end{table*}

To evaluate the proposed methodology, comprehensive evaluations were conducted across two distinct marine segmentation datasets, demonstrating its robustness and generalizability under varied operational paradigms :

\textbf{Oil Spill Drone} Dataset \cite{sels2024oilspill}: This dataset focuses on ecological monitoring with 847 high-resolution UAV images of resolution 1920×1080, of port environments containing oil spills, water surfaces, vessels, and shoreline structures. The dataset emphasizes fine boundary detection between spilt hydrocarbons and water surfaces, with per-pixel annotations verified by environmental scientists. A stratified 75-25 split considering the spatial distribution of spills to prevent overlapping regions between sets is implemented. The aerial perspective and thin oil sheens present unique challenges compared to USV-based imagery. This dataset is leveraged as the coastal RGB Earth observation dataset for this study.

Marine Semantic Segmentation Training Dataset (\textbf{MaSTr1325}) \cite{bovcon2019mastr}: This benchmark contains 1,325 images of resolution 512×384 captured from unmanned surface vehicles (USVs) operating in coastal environments over two years. The dataset spans five object categories critical for autonomous navigation: water, sky, vessels, obstacles, and animals. Following previous work \cite{bovcon2022wasr}, an 80-20 train-test split is employed with temporal stratification to prevent leakage between similar scenes captured in sequence. The maritime environments exhibit challenging conditions including variable illumination, wave reflections, and occlusions from spray/mist. This dataset is leveraged for the near-surface Earth observation of coastal conditions. 

Training was done using the NVIDIA TESLA P100, and inference was run using the NVIDIA 2080 and Intel 4-core XEON CPU. A batch size of 8 was used with the Adam optimizer \cite{kingma2014adam} for 300 epochs.

On investigating the role of the Residual Block Configuration
within LEMMA’s three-branch design, the authors found 7, 7, and 1 residual blocks (in the LFB, MFB, and HFB, respectively) for the MaSTr1325 dataset and 6, 7, and 4 blocks for the Oil Spill Dataset to offer an optimal balance between model complexity and segmentation accuracy. While adding more blocks marginally improves performance, it would also increase computational cost.

\begin{table}[t]
\renewcommand{\arraystretch}{1.3}
\centering
\small
\begin{tabular}{|l|c|c|}
\textbf{Model} & \textbf{mIoU} & \textbf{\#Params (M)} \\ \hline
\multicolumn{3}{|c|}{ResNet-50 (R) Backbones} \\ \hline
R-LinkNet \cite{chaurasia2017linknet} & 90.06 & - \\
R-SGDBNet \cite{dong2024sgdbnet} & 90.22 & - \\
R-MANet \cite{fan2020ma} & 91.92 & 26 \\
R-FPN \cite{lin2017fpn} & 91.95 & 28 \\
R-DeepLabv3 \cite{chen2017rethinking} & 92 & 39.6 \\
R-UNet \cite{colombo2020computer} & 92.04 & 24.5 \\
R-Segformer \cite{xie2021segformer} & 92.04 & 62.6 \\
R-DeepLabv3+ \cite{chen2018encoder} & 92.23 & 40 \\
R-UPerNet \cite{xiao2018unified} & 92.33 & 30 \\
R-GSSNet \cite{chen2023gss} & 92.66 & - \\
\hline
\multicolumn{3}{|c|}{Efficient-Net-b1 (E) Backbones} \\ \hline
E-UPerNet & 90.98 & - \\
E-LinkNet & 91.08 & 7.7 \\
E-SegFormer \cite{xie2021segformer} & 91.2 & 3.7 \\
E-DeepLabv3+ & 91.33 & 5 \\
E-DeepLabv3 & 91.65 & 5 \\
E-FPN & 91.71 & 7.7 \\
E-UNet & 91.87 & 7.7 \\
E-MANet & 91.26 & 7.7 \\
E-GSSNet & 92.14 & - \\
\hline
\textbf{LEMMA (Ours)} & \textbf{93.42} & \textbf{1.01} \\
\hline
\end{tabular}
\caption{Comparison of various models with LEMMA on mIoU across the Oil Spill Drone \cite{sels2024oilspill} dataset. Parameters are provided for both ResNet-50 (R) and Efficient-Net-b1 backbones (E) where applicable. Results for LEMMA are of residual block (Section \ref{sec: Method}) configurations 6,7,4. Missing data is due to unreported values in original works or lack of publicly available weights. Parameters for some of the models have been calculated by the parameter count of backbones and other modules of the architecture as made available by authors. }
\label{tab:Oil_Results}
\end{table}

\begin{figure*}[]
    \centering
    \includegraphics[width=\textwidth]{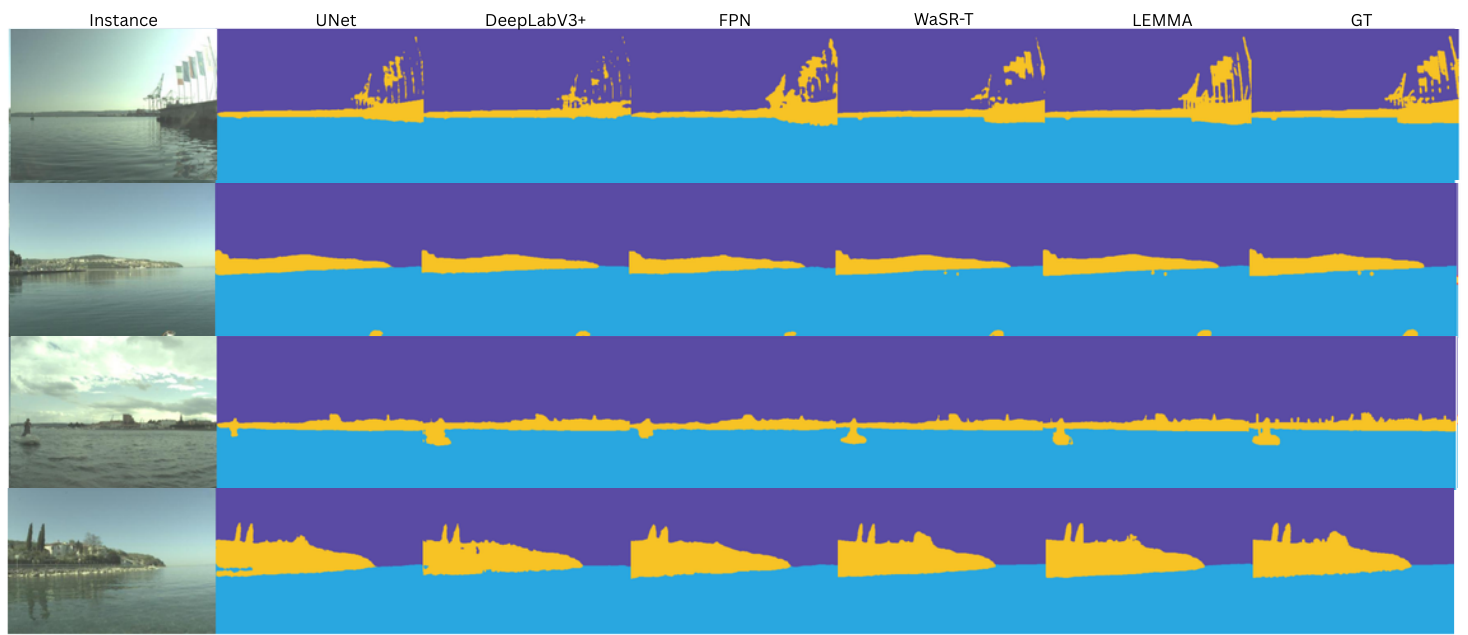}
    \caption{Visualization of the qualitative outputs on the  Mastr1325 Dataset.}
    \label{fig:mastr_qualitative}
\end{figure*}

\begin{figure*}[]
    \centering
    \includegraphics[width=\textwidth]{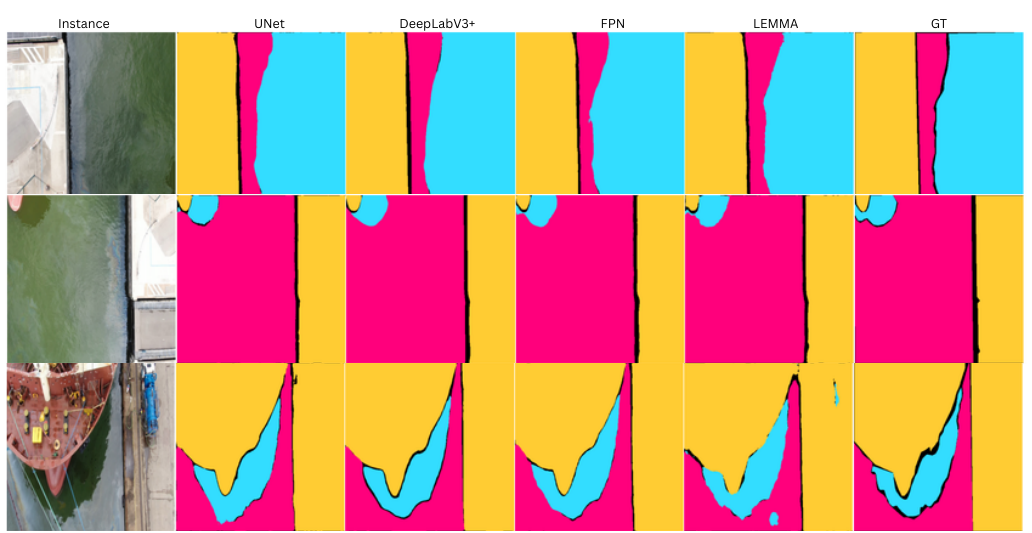}
    \caption{Visualization of the qualitative outputs on the Oil Spill Drone Dataset.}
    \label{fig:oil_qualitative}
\end{figure*}

Table \ref{tab:Mastr_results} and Table \ref{tab:Oil_Results} provide a consolidated view of the performance metrics: mean Intersection-over-Union (mIoU), trainable parameters, GFLOPs, and inference time. Comparisons are drawn against widely adopted CNN and Transformer-based approaches and specialised marine segmentation networks. Fig. \ref{fig:mastr_qualitative} and Fig. \ref{fig:oil_qualitative} offer qualitative results illustrating how LEMMA parses fine-scale structures across marine environments. 

As seen in Table \ref{tab:Mastr_results}, despite requiring up to 71× fewer trainable parameters than competing methods, LEMMA achieves comparable or better segmentation quality on the Mastr1325 dataset. In particular, mIoU often falls within a 1–2\% margin of heavier backbones with upto 88.5\% decrease in GFLOPs at 84.65\% lesser inference time. This is remarkable given that many of these larger models leverage massive pre-trained backbones (e.g., on ImageNet) or rely on Transformers for multi-level feature fusion. In contrast, LEMMA trains efficiently from scratch, leveraging the early-edge awareness inherent in Laplacian pyramids to bypass dense feature map computation, leading to around 1 Million parameters, with upto 17.83 GFLOPs and a fast inference time of 7.3 ms at its heaviest. 

WODIS \cite{chen2021wodis}, a neural network specifically designed for water obstacle detection on automatic surface vehicles in the marine environment, achieves a mIoU of 91.3\% at 43.2 FPS on the MaSTr1325 test set. However, this result is 7.67\% lower in mIoU and 68.47\% slower than the proposed method. Similarly, WaSR-T, a transformer-based network acheives 99.80\% mIoU, but requires about 71x more trainable parameters and 86.67\% more GFLOPs. Even lightweight methods for segmentation such as LETNet \cite{xu2023lightweight} and UISSNet \cite{he2024uiss} lack by 14-15\% mIoU and incurr a heavier cost of upto 44x.

Moving to Table \ref{tab:Oil_Results}, LEMMA demonstrates better performance than various SOTA methods with various backbone selections. Here, the superior performance of LEMMA, at a much lesser cost of upto 62x is noted. The reduced mIoU in Table \ref{tab:Oil_Results} as compared to Mastr1325 is primarily due to environmental factors such as reflections, waves, glare, and weather, which create inconsistencies in pixel intensities and impact the pyramidal representations, also visible in Fig. \ref{fig:oil_qualitative}. Legacy SOTA methods like DeepLabv3, DeepLabv3+, UNet and recent methods like the GSSNet and MANet are outperformed by LEMMA, which showcases an mIoU of 93.42\% at only 1.01M trainable parameters.

The most striking advantage of LEMMA is its ability to seamlessly adapt to contrasting viewpoints (surface level for MaSTr1325 vs. aerial for Oil Spill Drone). Many state-of-the-art Earth Observation segmentation methods focus heavily on a single domain or incorporate specialised attention mechanisms and multimodal inputs to improve results, often at the cost of significantly higher complexity. LEMMA, on the other hand, uses a consistent, pyramid-based feature extraction that excels at highlighting thin boundaries in both small USV obstacle segmentation (e.g., floating debris or buoys) and large-scale UAV oil spill delineation (e.g., faint hydrocarbon sheens on water), showing cross-platform robustness. The Laplacian pyramids implicitly suppress low-frequency illumination drift caused by factors such as sun glint, water reflectance, and haze.


Qualitative Analysis in Fig. \ref{fig:mastr_qualitative} and Fig. \ref{fig:oil_qualitative} showcase segmentation masks produced by LEMMA in both USV-based and UAV-based scenarios. The left column contains input images, followed by results of UNet \cite{ronneberger2015u}, DeepLabv3 \cite{chen2017rethinking}, FPN \cite{lin2017fpn}, WaSR-T \cite{zust2022temporal} and the proposed model, and the right column is the ground-truth mask. As seen across both datasets, the proposed methodology produces superior results for different frames of reference captured by aerial drones and surface-level sea vehicles. Additionally, LEMMA excels at segmenting not just oil spills but also obstacles like buoys, shorelines and piers at a much lesser computational load.


However, as noted in the last row of Fig. \ref{fig:oil_qualitative}, a failure case is demonstrated. The authors believe this is primarily due to the content of the images. A large portion of the RGB image in Row 3 is covered by the reflection of the ship. Hence, in its Laplacian pyramid, due to the reflection, the intensity values of the water and the reflection are very similar, causing edges to be blurred and to be not as prominent. This leads our edge-guided model to inaccurately segment areas. These limitations of LEMMA as visible in the mask when compared to those generated by pretraining-dependent, computationally heavy models.


\section{Ablation Study}
\label{Ablation}

In this section, we aim to evaluate the effects and analyze the trends observed through experiments involving varying residual blocks and loss functions for both datasets. This exhaustive experimentation justifies our choice of hyperparameters and enables our model to achieve state-of-the-art results with significantly fewer trainable parameters.
\begin{table}[]
\begin{tabular}{|c|c|c|c|c|}
\hline
\textbf{Dataset} & \textbf{NRB$_L$} & \textbf{NRB$_M$} & \textbf{NRB$_H$} & \textbf{mIOU} \\ \hline
\multirow{13}{*}{\textbf{Mastr1325}} & 3 & 3 & 3 & 0.9864 \\ 
 & 3 & 4 & 3 & 0.9871 \\ 
 & 3 & 5 & 3 & 0.9876 \\ 
 & 3 & 6 & 3 & 0.9876 \\ 
 & 3 & 7 & 3 & 0.9886 \\ \cline{2-5} 
 & 4 & 7 & 3 & 0.9882 \\ 
 & 5 & 7 & 3 & 0.9890 \\ 
 & 6 & 7 & 3 & 0.9894 \\ 
 & 7 & 7 & 3 & 0.9895 \\ \cline{2-5} 
 & \textbf{7} & \textbf{7} & \textbf{1} & \textbf{0.9896} \\ 
 & 7 & 7 & 2 & 0.9893 \\ 
 & 7 & 7 & 3 & 0.9892 \\ 
 & 7 & 7 & 4 & 0.9894 \\ \hline
\multirow{13}{*}{\textbf{Oil Spill}} & 3 & 3 & 3 & 0.9063 \\ 
 & 3 & 4 & 3 & 0.9148 \\ 
 & 3 & 5 & 3 & 0.9055 \\ 
 & 3 & 6 & 3 & 0.9132 \\ 
 & 3 & 7 & 3 & 0.9151 \\ \cline{2-5} 
 & 4 & 7 & 3 & 0.9188 \\ 
 & 5 & 7 & 3 & 0.9149 \\ 
 & 6 & 7 & 3 & 0.9294 \\
 & 7 & 7 & 3 & 0.9191 \\ \cline{2-5} 
 & 6 & 7 & 1 & 0.9299 \\ 
 & 6 & 7 & 2 & 0.9217 \\ 
 & \textbf{6} & \textbf{7} & \textbf{4} & \textbf{0.9342} \\
  \hline 
\end{tabular}
\caption{Impact of Number of residual blocks in each branch of the model and the corresponding results on both datasets. The best results for every branch are bolded.}
\label{tab:NRB-ablation}
\end{table}

Table \ref{tab:NRB-ablation} illustrates the impact of the number of residual blocks in each branch of the model. As mentioned in Section \ref{sec: Method}, the number of residual blocks in the lower, middle, and higher branches are denoted as NRB$_L$, NRB$_M$ and NRB$_H$, respectively. For optimal performance, the ideal configuration of residual blocks selected for the MaSTr1325 dataset is 7, 7, 1, whereas for the Oil Spill dataset, it is 6, 7, 4. For both datasets, the optimal residual block configuration is determined by keeping the number of blocks in two branches fixed while varying the other. The highest mIoU of 98.96\% is obtained with the 7, 7, 1 configuration for the MaSTr1325 dataset and 93.42\% is obtained with 6, 7, 4 on the Oil Spill dataset. Increasing the number of residual blocks in the top branch slightly reduces performance while increasing the number of parameters and GFLOPs, as it requires additional computations on larger feature maps. Similarly, the 6, 7, and 4 configurations are chosen for the Oil Spill dataset, where increasing NRB$_L$ beyond 6 blocks slightly degrades performance, and increasing NRB$_H$ up to 4 results in diminishing returns. This demonstrates that carefully chosen configurations can achieve high mIoU without unnecessarily increasing complexity.
\begin{table}[]
\centering
\begin{tabular}{|c|c|c|}
\hline
\textbf{Dataset} & \textbf{Loss Function} & \textbf{mIOU} \\ \hline
\multirow{3}{*}{\textbf{MaSTr1325}} & \textbf{Focal} & \textbf{0.9897} \\ 
 & Dice & 0.9872 \\ 
 & CE+Dice & 0.9886 \\ \hline
\multirow{3}{*}{\textbf{Oil Spill}} & \textbf{Focal} & \textbf{0.9342} \\ 
 & Dice & 0.9262 \\ 
 & CE+Dice & 0.9294 \\ \hline
\end{tabular}
\caption{Impact of various loss functions with best configurations of residual blocks on both datasets. The best loss function for each dataset is bolded.}
\label{tab:Loss-ablation}
\end{table}

Table \ref{tab:Loss-ablation} evaluates the effect of various loss functions. For these experiments, commonly used segmentation loss functions are considered, namely Focal Loss \cite{lin2017focal}, Dice Loss \cite{milletari2016v}, and Cross Entropy + Dice Loss. The best performance on the Mastr1325 dataset is observed with Focal Loss, achieving an mIoU of 98.97\%, while the Oil Spill dataset achieves its highest performance of 93.42\% using Focal Loss as well.

\section{Conclusion and Future Work}

In this work, we introduced LEMMA, a novel lightweight semantic segmentation network for marine segmentation which bypasses heavy computational loads by employing Laplacian Pyramids for feature extraction along with multi-level processing with residual blocks. LEMMA demonstrates comparable performance with up to 71x fewer parameters, making it deployable in real-world scenarios which need quick segmentation of obstacles and oil spills.

While LEMMA demonstrates promising results with significantly reduced parameter counts, several limitations remain. First, while the current datasets are standard in the marine domain, they are limited in size and diversity. Limitations of this method include failure cases when reflections, waves, glare and environmental conditions are encountered. These affect the Laplacian pyramid and hence the subsequent construction of an accurate mask.
Finally, LEMMA's current architecture uses fixed pyramid levels and static residual block configurations; future extensions will explore adaptive pyramidal decomposition and dynamic depth allocation based on image content, aiming to further improve efficiency-accuracy trade-offs.

\section{Acknowledgement}\
\label{Ack}

We would like to thank Mars Rover Manipal, an interdisciplinary student project of MAHE, for providing the essential resources and infrastructure that supported our research. We also extend our gratitude to Mohammed Sulaiman for his contributions in facilitating access to additional resources crucial to this work.
{
    \small
    \bibliographystyle{ieeenat_fullname}
    \bibliography{main}

@String(CVPR= {IEEE Conf. Comput. Vis. Pattern Recog.})

@String(ECCV= {Eur. Conf. Comput. Vis.})

@String(TOG= {ACM Trans. Graph.})

@String(ICIP = {IEEE Int. Conf. Image Process.})

@String(CVPR  = {CVPR})

@String(ECCV  = {ECCV})

@String(TOG   = {ACM TOG})

@String(ICIP  = {ICIP})

@inproceedings{long2015fully,
  title={Fully convolutional networks for semantic segmentation},
  author={Long, Jonathan and Shelhamer, Evan and Darrell, Trevor},
  booktitle={Proceedings of the IEEE conference on computer vision and pattern recognition},
  pages={3431--3440},
  year={2015}
}

@article{fiscella2000oil,
  title={Oil spill detection using marine SAR images},
  author={Fiscella, B and Giancaspro, A and Nirchio, F and Pavese, P and Trivero, Paolo},
  journal={International Journal of Remote Sensing},
  volume={21},
  number={18},
  pages={3561--3566},
  year={2000},
  publisher={Taylor \& Francis}
}

@inproceedings{xie2015holistically,
  title={Holistically-nested edge detection},
  author={Xie, Saining and Tu, Zhuowen},
  booktitle={Proceedings of the IEEE international conference on computer vision},
  pages={1395--1403},
  year={2015}
}

@incollection{burt1987laplacian,
  title={The Laplacian pyramid as a compact image code},
  author={Burt, Peter J and Adelson, Edward H},
  booktitle={Readings in computer vision},
  pages={671--679},
  year={1987},
  publisher={Elsevier}
}

@article{lee2023semantic,
  title={Semantic Segmentation Network Slimming and Edge Deployment for Real-Time Forest Fire or Flood Monitoring Systems Using Unmanned Aerial Vehicles},
  author={Lee, Youn Joo and Jung, Ho Gi and Suhr, Jae Kyu},
  journal={Electronics},
  volume={12},
  number={23},
  pages={4795},
  year={2023},
  publisher={MDPI}
}

@article{paris2011local,
  author    = {S. Paris and S. W. Hasinoff and J. Kautz},
  title     = {Local Laplacian Filters: Edge-aware Image Processing with a Laplacian Pyramid},
  journal   = {ACM Transactions on Graphics (TOG)},
  volume    = {30},
  number    = {4},
  pages     = {1-12},
  year      = {2011},
  doi       = {10.1145/2010324.1964963},
  publisher = {ACM}
}

@article{bovcon2022wasr,
  author    = {B. Bovcon and M. Kristan},
  title     = {WaSR—A Water Segmentation and Refinement Maritime Obstacle Detection Network},
  journal   = {IEEE Transactions on Cybernetics},
  volume    = {52},
  number    = {12},
  pages     = {12661–12674},
  year      = {2022},
  doi       = {10.1109/TCYB.2022.3159784},
  publisher = {IEEE}
}

@article{zhang2024bems,
  author    = {L. Zhang and X. Sun and Z. Li and D. Kong and J. Liu and P. Ni},
  title     = {Boundary Enhancement-Driven Accurate Semantic Segmentation Networks for Unmanned Surface Vessels in Complex Marine Environments},
  journal   = {IEEE Sensors Journal},
  volume    = {24},
  number    = {15},
  pages     = {24972–24987},
  year      = {2024},
  doi       = {10.1109/JSEN.2024.3312492},
  publisher = {IEEE}
}

@article{feng2024dual,
  author    = {H. Feng and W. Liu and H. Xu and J. He},
  title     = {A Lightweight Dual-Branch Semantic Segmentation Network for Enhanced Obstacle Detection in Ship Navigation},
  journal   = {Engineering Applications of Artificial Intelligence},
  volume    = {127},
  pages     = {105049},
  year      = {2024},
  doi       = {10.1016/j.engappai.2024.105049},
  publisher = {Elsevier}
}

@inproceedings{bovcon2019mastr,
  author    = {B. Bovcon and J. Muhovic and J. Pers and M. Kristan},
  title     = {The MASTR1325 Dataset for Training Deep USV Obstacle Detection Models},
  booktitle = {2019 IEEE/RSJ International Conference on Intelligent Robots and Systems (IROS)},
  pages     = {3431–3438},
  year      = {2019},
  doi       = {10.1109/IROS40897.2019.8968061},
  publisher = {IEEE}
}

@article{sels2024oilspill,
  author    = {S. Sels and S. Vanlanduit and T. D. Kerf},
  title     = {Annotated RGB Images of Oil Spills in a Port Environment},
  journal   = {Remote Sensing},
  volume    = {16},
  number    = {2},
  pages     = {312},
  year      = {2024},
  doi       = {10.3390/rs16020312},
  publisher = {MDPI}
}

@article{denton2015deep,
  title={Deep generative image models using a laplacian pyramid of adversarial networks},
  author={Denton, Emily L and Chintala, Soumith and Fergus, Rob and others},
  journal={Advances in neural information processing systems},
  volume={28},
  year={2015}
}

@InProceedings{10.1007/978-3-319-46487-9_32,
author="Ghiasi, Golnaz
and Fowlkes, Charless C.",
editor="Leibe, Bastian
and Matas, Jiri
and Sebe, Nicu
and Welling, Max",
title="Laplacian Pyramid Reconstruction and Refinement for Semantic Segmentation",
booktitle="Computer Vision -- ECCV 2016",
year="2016",
publisher="Springer International Publishing",
address="Cham",
pages="519--534",
isbn="978-3-319-46487-9"
}

@article{kasliwal2024lapgsr,
  title={LapGSR: Laplacian Reconstructive Network for Guided Thermal Super-Resolution},
  author={Kasliwal, Aditya and Gakhar, Ishaan and Kamani, Aryan and Seth, Pratinav and Verma, Ujjwal},
  journal={arXiv preprint arXiv:2411.07750},
  year={2024}
}

@inproceedings{lin2021drafting,
  title={Drafting and revision: Laplacian pyramid network for fast high-quality artistic style transfer},
  author={Lin, Tianwei and Ma, Zhuoqi and Li, Fu and He, Dongliang and Li, Xin and Ding, Errui and Wang, Nannan and Li, Jie and Gao, Xinbo},
  booktitle={Proceedings of the IEEE/CVF conference on computer vision and pattern recognition},
  pages={5141--5150},
  year={2021}
}

@inproceedings{chen2022water,
  title={Water Segmentation with Superior Guidance and Aligned Fusion Strategy for Unmanned Surface Vehicles in Maritime Environment},
  author={Chen, Zhengquan and Dong, Ding and Zhao, Binyu and Zhang, Wei},
  booktitle={2022 IEEE International Conference on Unmanned Systems (ICUS)},
  pages={1252--1257},
  year={2022},
  organization={IEEE}
}

@inproceedings{taipalmaa2019high,
  title={High-resolution water segmentation for autonomous unmanned surface vehicles: A novel dataset and evaluation},
  author={Taipalmaa, Jussi and Passalis, Nikolaos and Zhang, Honglei and Gabbouj, Moncef and Raitoharju, Jenni},
  booktitle={2019 IEEE 29th International Workshop on Machine Learning for Signal Processing (MLSP)},
  pages={1--6},
  year={2019},
  organization={IEEE}
}

@inproceedings{yao2021shorelinenet,
  title={ShorelineNet: An efficient deep learning approach for shoreline semantic segmentation for unmanned surface vehicles},
  author={Yao, Linghong and Kanoulas, Dimitrios and Ji, Ze and Liu, Yuanchang},
  booktitle={2021 IEEE/RSJ International Conference on Intelligent Robots and Systems (IROS)},
  pages={5403--5409},
  year={2021},
  organization={IEEE}
}

@inproceedings{zhao2017pyramid,
  title={Pyramid scene parsing network},
  author={Zhao, Hengshuang and Shi, Jianping and Qi, Xiaojuan and Wang, Xiaogang and Jia, Jiaya},
  booktitle={Proceedings of the IEEE conference on computer vision and pattern recognition},
  pages={2881--2890},
  year={2017}
}

@article{chen2017rethinking,
  title={Rethinking atrous convolution for semantic image segmentation},
  author={Chen, Liang-Chieh and Papandreou, George and Schroff, Florian and Adam, Hartwig},
  journal={arXiv preprint arXiv:1706.05587},
  year={2017}
}

@inproceedings{sun2019deep,
  title={Deep high-resolution representation learning for human pose estimation},
  author={Sun, Ke and Xiao, Bin and Liu, Dong and Wang, Jingdong},
  booktitle={Proceedings of the IEEE/CVF conference on computer vision and pattern recognition},
  pages={5693--5703},
  year={2019}
}

@inproceedings{ronneberger2015u,
  title={U-net: Convolutional networks for biomedical image segmentation},
  author={Ronneberger, Olaf and Fischer, Philipp and Brox, Thomas},
  booktitle={Medical image computing and computer-assisted intervention--MICCAI 2015: 18th international conference, Munich, Germany, October 5-9, 2015, proceedings, part III 18},
  pages={234--241},
  year={2015},
  organization={Springer}
}

@inproceedings{zust2022temporal,
  title={Temporal Context for Robust Maritime Obstacle Detection. In 2022 IEEE},
  author={Zust, Lojze and Kristan, Matej},
  booktitle={RJS International Conference on Intelligent Robots and Systems (IROS)},
  volume={2},
  number={3},
  pages={5},
  year={2022}
}

@article{cao2024bemrf,
  title={BEMRF-Net: Boundary enhancement and multiscale refinement fusion for building extraction from remote sensing imagery},
  author={Cao, Shaohan and Feng, Dejun and Liu, Suning and Xu, Wanqi and Chen, Hongyu and Xie, Yakun and Zhang, Heng and Pirasteh, Saied and Zhu, Jun},
  journal={IEEE Journal of Selected Topics in Applied Earth Observations and Remote Sensing},
  year={2024},
  publisher={IEEE}
}

@inproceedings{lin2017focal,
  title={Focal loss for dense object detection},
  author={Lin, Tsung-Yi and Goyal, Priya and Girshick, Ross and He, Kaiming and Doll{\'a}r, Piotr},
  booktitle={Proceedings of the IEEE international conference on computer vision},
  pages={2980--2988},
  year={2017}
}

@inproceedings{milletari2016v,
  title={V-net: Fully convolutional neural networks for volumetric medical image segmentation},
  author={Milletari, Fausto and Navab, Nassir and Ahmadi, Seyed-Ahmad},
  booktitle={2016 fourth international conference on 3D vision (3DV)},
  pages={565--571},
  year={2016},
  organization={Ieee}
}

@article{badrinarayanan2017segnet,
  title={Segnet: A deep convolutional encoder-decoder architecture for image segmentation},
  author={Badrinarayanan, Vijay and Kendall, Alex and Cipolla, Roberto},
  journal={IEEE transactions on pattern analysis and machine intelligence},
  volume={39},
  number={12},
  pages={2481--2495},
  year={2017},
  publisher={IEEE}
}

@phdthesis{abdollahi2022automatic,
  title={Automatic updating and verification of road maps using high-resolution remote sensing images based on advanced machine learning methods},
  author={Abdollahi, Abolfazl},
  year={2022},
  school={University of Technology Sydney (Australia)}
}

@article{thisanke2023semantic,
  title={Semantic segmentation using Vision Transformers: A survey},
  author={Thisanke, Hans and Deshan, Chamli and Chamith, Kavindu and Seneviratne, Sachith and Vidanaarachchi, Rajith and Herath, Damayanthi},
  journal={Engineering Applications of Artificial Intelligence},
  volume={126},
  pages={106669},
  year={2023},
  publisher={Elsevier}
}

@article{manakitsa2024review,
  title={A review of machine learning and deep learning for object detection, semantic segmentation, and human action recognition in machine and robotic vision},
  author={Manakitsa, Nikoleta and Maraslidis, George S and Moysis, Lazaros and Fragulis, George F},
  journal={Technologies},
  volume={12},
  number={2},
  pages={15},
  year={2024},
  publisher={MDPI}
}

@article{wang2022underwater,
  author    = {Wang, J. and He, X. and Shao, F. and Lu, G. and Hu, R. and Jiang, Q.},
  title     = {Semantic Segmentation Method of Underwater Images Based on Encoder-Decoder Architecture},
  journal   = {PLoS ONE},
  volume    = {17},
  number    = {8},
  pages     = {e0272666},
  year      = {2022},
  doi       = {10.1371/journal.pone.0272666},
  publisher = {Public Library of Science}
}

@article{dai2025ldanet,
  title={LDANet: enhancing USV's capacity for better segmentation of complex waterway scenes},
  author={Dai, Tongyang and Xiang, Huiyu and Leng, Chongjie and Huang, Song and He, Guanghui and Han, Shishuo},
  journal={Soft Computing},
  pages={1--15},
  year={2025},
  publisher={Springer}
}

@inproceedings{strudel2021segmenter,
  title={Segmenter: Transformer for semantic segmentation},
  author={Strudel, Robin and Garcia, Ricardo and Laptev, Ivan and Schmid, Cordelia},
  booktitle={Proceedings of the IEEE/CVF international conference on computer vision},
  pages={7262--7272},
  year={2021}
}

@article{chen2021wodis,
  title={WODIS: Water obstacle detection network based on image segmentation for autonomous surface vehicles in maritime environments},
  author={Chen, Xiang and Liu, Yuanchang and Achuthan, Kamalasudhan},
  journal={IEEE Transactions on Instrumentation and Measurement},
  volume={70},
  pages={1--13},
  year={2021},
  publisher={IEEE}
}

@article{poudel2019fast,
  title={Fast-scnn: Fast semantic segmentation network},
  author={Poudel, Rudra PK and Liwicki, Stephan and Cipolla, Roberto},
  journal={arXiv preprint arXiv:1902.04502},
  year={2019}
}

@article{pan2022deep,
  title={Deep dual-resolution networks for real-time and accurate semantic segmentation of traffic scenes},
  author={Pan, Huihui and Hong, Yuanduo and Sun, Weichao and Jia, Yisong},
  journal={IEEE Transactions on Intelligent Transportation Systems},
  volume={24},
  number={3},
  pages={3448--3460},
  year={2022},
  publisher={IEEE}
}

@article{simonyan2014very,
  title={Very deep convolutional networks for large-scale image recognition},
  author={Simonyan, Karen and Zisserman, Andrew},
  journal={arXiv preprint arXiv:1409.1556},
  year={2014}
}

@inproceedings{chen2018encoder,
  title={Encoder-decoder with atrous separable convolution for semantic image segmentation},
  author={Chen, Liang-Chieh and Zhu, Yukun and Papandreou, George and Schroff, Florian and Adam, Hartwig},
  booktitle={Proceedings of the European conference on computer vision (ECCV)},
  pages={801--818},
  year={2018}
}

@inproceedings{chen2022mobile,
  title={Mobile-former: Bridging mobilenet and transformer},
  author={Chen, Yinpeng and Dai, Xiyang and Chen, Dongdong and Liu, Mengchen and Dong, Xiaoyi and Yuan, Lu and Liu, Zicheng},
  booktitle={Proceedings of the IEEE/CVF conference on computer vision and pattern recognition},
  pages={5270--5279},
  year={2022}
}

@article{romera2017erfnet,
  title={Erfnet: Efficient residual factorized convnet for real-time semantic segmentation},
  author={Romera, Eduardo and Alvarez, Jos{\'e} M and Bergasa, Luis M and Arroyo, Roberto},
  journal={IEEE Transactions on Intelligent Transportation Systems},
  volume={19},
  number={1},
  pages={263--272},
  year={2017},
  publisher={IEEE}
}

@inproceedings{mehta2019espnetv2,
  title={Espnetv2: A light-weight, power efficient, and general purpose convolutional neural network},
  author={Mehta, Sachin and Rastegari, Mohammad and Shapiro, Linda and Hajishirzi, Hannaneh},
  booktitle={Proceedings of the IEEE/CVF conference on computer vision and pattern recognition},
  pages={9190--9200},
  year={2019}
}

@article{li2019dabnet,
  title={Dabnet: Depth-wise asymmetric bottleneck for real-time semantic segmentation},
  author={Li, Gen and Yun, Inyoung and Kim, Jonghyun and Kim, Joongkyu},
  journal={arXiv preprint arXiv:1907.11357},
  year={2019}
}

@inproceedings{lou2021cfpnet,
  title={Cfpnet: Channel-wise feature pyramid for real-time semantic segmentation},
  author={Lou, Ange and Loew, Murray},
  booktitle={2021 IEEE international conference on image processing (ICIP)},
  pages={1894--1898},
  year={2021},
  organization={IEEE}
}

@article{xu2023lightweight,
  title={Lightweight real-time semantic segmentation network with efficient transformer and CNN},
  author={Xu, Guoan and Li, Juncheng and Gao, Guangwei and Lu, Huimin and Yang, Jian and Yue, Dong},
  journal={IEEE Transactions on Intelligent Transportation Systems},
  volume={24},
  number={12},
  pages={15897--15906},
  year={2023},
  publisher={IEEE}
}

@article{he2024uiss,
  title={UISS-Net: Underwater Image Semantic Segmentation Network for improving boundary segmentation accuracy of underwater images},
  author={He, ZhiQian and Cao, LiJie and Luo, JiaLu and Xu, XiaoQing and Tang, JiaYi and Xu, JianHao and Xu, GengYan and Chen, ZiWen},
  journal={Aquaculture International},
  volume={32},
  number={5},
  pages={5625--5638},
  year={2024},
  publisher={Springer}
}

@article{liu2025lgcgnet,
  title={LGCGNet: A local-global context guided network for real-time water surface semantic segmentation: T. Liu et al.},
  author={Liu, Ting and Luo, Peiqi and Wang, Guofeng and Zhang, Yuxin and Lu, Xiangyi and Dong, Mengyu},
  journal={Applied Intelligence},
  volume={55},
  number={7},
  pages={448},
  year={2025},
  publisher={Springer}
}

@inproceedings{colombo2020computer,
  author    = {Colombo, Mattia and Dolhasz, Alan and Harvey, Carlo},
  title     = {A Computer Vision-Inspired Automatic Acoustic Material Tagging System for Virtual Environments},
  booktitle = {Proceedings of the 2020 IEEE Conference on Games (CoG)},
  pages     = {736--739},
  year      = {2020}
}

@article{fan2020ma,
  author  = {Fan, Tongle and Wang, Guanglei and Li, Yan and Wang, Hongrui},
  title   = {{MA-Net: A Multi-Scale Attention Network for Liver and Tumor Segmentation}},
  journal = {IEEE Access},
  volume  = {8},
  pages   = {179650--179660},
  year    = {2020}
}

@article{chaurasia2017linknet,
  author  = {Chaurasia, Abhishek and Culurciello, Eugenio},
  title   = {{LinkNet: Exploiting Encoder Representations for Efficient Semantic Segmentation}},
  journal = {arXiv preprint arXiv:1707.03718},
  year    = {2017}
}

@inproceedings{lin2017fpn,
  author    = {Lin, Tsung-Yi and Doll{\'a}r, Piotr and Girshick, Ross and He, Kaiming and Hariharan, Bharath and Belongie, Serge},
  title     = {Feature Pyramid Networks for Object Detection},
  booktitle = {Proc. of the IEEE Conf. on Computer Vision and Pattern Recognition (CVPR)},
  pages     = {2117--2125},
  year      = {2017}
}

@inproceedings{xiao2018unified,
  author    = {Xiao, Tete and Liu, Yingcheng and Zhou, Bolei and Jiang, Yuning and Sun, Jian},
  title     = {Unified Perceptual Parsing for Scene Understanding},
  booktitle = {Proceedings of the European Conf. on Computer Vision (ECCV)},
  pages     = {418--434},
  year      = {2018}
}

@inproceedings{xie2021segformer,
  author    = {Xie, Enze and Wang, Wenhai and Yu, Zhiding and Anandkumar, Anima and Alvarez, Jose M. and Luo, Ping},
  title     = {SegFormer: Simple and Efficient Design for Semantic Segmentation with Transformers},
  booktitle = {Advances in Neural Information Processing Systems (NeurIPS)},
  year      = {2021}
}

@article{dong2024sgdbnet,
  title={SGDBNet: A scene-class guided dual branch network for port UAV images oil spill detection},
  author={Dong, Shaokang and Feng, Jiangfan},
  journal={Marine Pollution Bulletin},
  volume={208},
  pages={117019},
  year={2024},
  publisher={Elsevier},
  doi={10.1016/j.marpolbul.2024.117019}
}

@article{chen2023gss,
  title={Generative Semantic Segmentation},
  author={Chen, Jiaqi and Lu, Jiachen and Zhu, Xiatian and Zhang, Li},
  journal={Proceedings of the IEEE/CVF Conference on Computer Vision and Pattern Recognition (CVPR)},
  pages={12345--12354},
  year={2023},
  doi={10.1109/CVPR45657.2023.01234}
}

@article{kingma2014adam,
  title={Adam: A method for stochastic optimization},
  author={Kingma, Diederik P},
  journal={arXiv preprint arXiv:1412.6980},
  year={2014}
}

@article{temitope2020advances,
  title={Advances in remote sensing technology, machine learning and deep learning for marine oil spill detection, prediction and vulnerability assessment},
  author={Temitope Yekeen, Shamsudeen and Balogun, Abdul-Lateef},
  journal={Remote Sensing},
  volume={12},
  number={20},
  pages={3416},
  year={2020},
  publisher={MDPI}
}

@article{miao2018automatic,
  title={Automatic water-body segmentation from high-resolution satellite images via deep networks},
  author={Miao, Ziming and Fu, Kun and Sun, Hao and Sun, Xian and Yan, Menglong},
  journal={IEEE geoscience and remote sensing letters},
  volume={15},
  number={4},
  pages={602--606},
  year={2018},
  publisher={IEEE}
}

@article{zhou2019underwater,
  title={Underwater scene segmentation by deep neural network},
  author={Zhou, Yang and Wang, Jiangtao and Li, Baihua and Meng, Qinggang and Rocco, Emanuele and Saiani, Andrea},
  year={2019},
  publisher={Loughborough University}
}
}

\end{document}